\documentclass[manuscript,screen]{acmart}
\usepackage[ruled,vlined]{algorithm2e}

\SetKwInput{KwInput}{Input}
\SetKwInput{KwOutput}{Output}
\usepackage{braket}
\usepackage{tabularx}
\usepackage[normalem]{ulem}
\AtBeginDocument{%
  }

\setcopyright{acmlicensed}
\copyrightyear{2025}
\acmYear{2025}
\acmDOI{XXXXXXX.XXXXXXX}
\acmISBN{978-1-4503-XXXX-X/2018/06}

\begin{document}

\title[Gate-Based QRC for MTS on NISQ]{Multivariate Time Series Forecasting with Gate-Based Quantum Reservoir Computing on NISQ Hardware}

\author{Wissal Hamhoum}
\email{wissal.hamhoum@polymtl.ca}
\orcid{https://orcid.org/0009-0007-2813-5133}
\author{Soumaya Cherkaoui}
\email{soumaya.cherkaoui@polymtl.ca}
\affiliation{%
  \institution{Polytechnique Montréal}
  \city{Montreal}
  \country{Canada}
}

\author{Jean-Frédéric Laprade}
\affiliation{%
  \institution{Institut quantique, Université de Sherbrooke}
  \city{Sherbrooke}
  \country{Canada}}
\email{jean-frederic.laprade@usherbrooke.ca}

\author{Ola Ahmad}
\affiliation{%
  \institution{Thales cortAIx Labs}
  \city{Montréal}
  \country{Canada}
}
\email{ola.ahmad@thalesdigital.io}

\author{Shengrui Wang}
\affiliation{%
 \institution{Université de Sherbrooke}
 \city{Sherbrooke}
 \country{Canada}}
\email{shengrui.wang@usherbrooke.ca}





\renewcommand{\shortauthors}{Hamhoum et al.}

\begin{abstract}

Quantum reservoir computing (QRC) offers a hardware-friendly approach to temporal learning, yet most studies target univariate signals and overlook near-term hardware constraints. This work introduces a gate-based QRC for multivariate time series (MTS-QRC) that pairs injection and memory qubits and uses a Trotterized nearest-neighbor transverse-field Ising evolution optimized for current device connectivity and depth. On Lorenz-63 and ENSO, the method achieves a mean square error (MSE) of 0.0087 and 0.0036, respectively, performing on par with classical reservoir computing on Lorenz and above learned RNNs on both, while NVAR and clustered ESN remain stronger in some settings. On IBM Heron R2, MTS-QRC sustains accuracy with realistic depths and, interestingly, outperforms a noiseless simulator on ENSO; singular value analysis indicates that device noise can concentrate variance in feature directions, acting as an implicit regularizer for linear readout in this regime. These findings support the practicality of gate-based QRC for MTS forecasting on NISQ hardware and motivate systematic studies on when and how hardware noise benefits QRC readouts.
\end{abstract}




\maketitle

\section{Introduction}
\label{sec:introduction}
Time Series (TS) forecasting plays a critical role in a wide range of real-world applications, including weather prediction, financial modeling, control of cyber-physical systems, and network anomaly detection \cite{10583885,landauer_review_2025}. In most domains, the data originates from multivariate and highly nonlinear dynamical systems whose behavior is often characterized by chaos-extreme sensitivity to initial conditions, low long-term predictability, and complex temporal dependencies \cite{tian2020chaotic}. These characteristics make forecasting particularly challenging, even for advanced machine learning models. Capturing the intricate structure of multivariate chaotic time series requires models to be both expressive and memory-efficient while being robust to noise and adaptable to varying temporal patterns.

Among the many ML approaches designed for sequential data, Reservoir Computing (RC) has emerged as an efficient and powerful paradigm \cite{yan_emerging_2024}. RC models rely on a three-layer architecture in which only the output layer is trained. The core of the system, the reservoir layer, projects the input data into a high-dimensional dynamical space through fixed, nonlinear transformations. This setup enables rich temporal feature extraction at a reduced computational cost. The reservoir layer can be implemented using conventional neural networks, as in Echo State Networks (ESN) \cite{9966815} and Liquid State Machines (LSM) \cite{wijesinghe2019analysis}, or via physical systems exhibiting rich intrinsic dynamics, such as mechanical oscillators \cite{coulombe2017computing} or photonic systems \cite{vandoorne2014experimental}.

Recently, quantum systems have gained attention as promising candidates for the realization of physical reservoirs ~\cite{kornjaca_large-scale_2024}. Thanks to their unique quantum properties, such as superposition and entanglement, quantum reservoirs can produce rich, high-dimensional feature maps that are difficult to achieve classically, in addition to a memory retention capacity\cite{gotting_exploring_2023}. This led to the emergence of Quantum Reservoir Computing (QRC), a framework within Quantum Machine Learning (QML) that holds strong potential in sequential data processing. Additionally, QRC stands out as one of the few QML models well-suited for implementation on Noisy Intermediate-Scale Quantum (NISQ) devices \cite{fujii2021quantum}. Unlike variational quantum algorithms, QRC does not rely on trainable quantum circuits, thereby avoiding issues such as barren plateaus and complex optimization landscapes \cite{mcclean2018barren}. Moreover, studies showed that in some cases, noise can be beneficial for QRC by enriching the dynamics of the system \cite{domingo_taking_2023}.

Motivated by its theoretical potential and the rapid advancement of quantum technologies, several works have explored the implementation of QRC across various quantum platforms. This includes bosonic systems \cite{mujal2021quantum},  fermionic systems \cite{ghosh2019quantum}, neutral atom platforms \cite{kornjaca_large-scale_2024}, and gate-based quantum processors \cite{yasuda2023quantum}. Each platform naturally produces distinct quantum dynamics, offering different computational potentials. In particular, digital quantum computers, also known as gate-based computers, offer greater flexibility and programmability compared to their analog counterparts that are limited to specific interactions or fixed dynamics. This flexibility facilitates the emulation of a wide range of QRC dynamics, from Hamiltonian-based evolutions under various interaction models \cite{sakurai2025simplehamiltoniandynamicspowerful,sasaki2025hamiltoniandrivenarchitecturesnonmarkovianquantum} to non-Hamiltonian approaches based on either structurally random circuits or parameterized quantum circuits initialized with randomly sampled parameters~\cite{ahmed2405prediction}. Furthermore, digital quantum platforms benefit from better-developed error mitigation and correction methods, making them strong candidates for a QRC on near-term hardware that yields meaningful results. 

Despite the potential, most studies focus on univariate TS, and only a few apply QRC to Multivariate TS (MTS) \cite{xia_configured_2023,steinegger2025predicting} which are closer to real application scenarios. Furthermore, to the best of our knowledge, none of them has considered the real quantum hardware constraints in their designs. Therefore, in this work, we propose an implementation of a Hamiltonian-based QRC that addresses the task of MTS forecasting with a design tailored for an efficient execution on a gate-based quantum computer.  We benchmark our model on famous chaotic systems: Lorenz-63 and El Niño Southern Oscillations (ENSO), both of which involve multiple interacting variables under chaotic dynamics. Beyond benchmarking, we report an intriguing observation: for ENSO series, our hardware execution of MTS-QRC yields consistently better results than the simulation. While a deep investigation of this phenomenon is needed, we analyzed potential pathways by which hardware noise contributes to improved results. Notably, these results are obtained without the use of error mitigation, suggesting that the hardware noise itself is contributing to the results.
The main contributions of this paper are as follows:
\begin{itemize}
\item We propose a new extension of the QRC framework to MTS forecasting, termed MTS-QRC, which leverages a gate-based quantum reservoir to process multidimensional time series data.
\item We benchmark our model against similar classical approaches and show that it achieves competitive results.
\item We implement and evaluate the proposed MTS-QRC model on real quantum hardware, specifically the IBM Heron R2 device, thereby demonstrating its practical feasibility beyond classical simulation.

\end{itemize}

The rest of this paper is organized as follows. Section~\ref{sec:background} briefly outlines the principles of reservoir computing and introduce the fundamentals of quantum reservoirs. Section~\ref{sec:system_model} details our MTS-QRC model, including the quantum circuit design, encoding schemes, and information retrieval protocol. Section~\ref{sec:experiments} presents the experimental setup, datasets, and evaluation metrics. In section Section~\ref{sec:results}, we provide the experimental results and discussions.  Finally, the conclusion is given in Section~\ref{sec:conclusion}.

\section{Related works}
\label{sec:related_works}
Reservoir Computing (RC) has been widely adopted for time series forecasting because of its simplicity and efficiency in modeling nonlinear dynamics \cite{shahi_prediction_2022}. In particular, Echo State Networks (ESNs) have demonstrated strong capabilities in capturing complex temporal dynamics in financial data, as shown in \cite{dos2025reservoir}. In the field of biology, the authors of \cite{huang2025study} utilized an MTS ESN to model highly non-linear relationships in the time series of the penicillin fermentation process.
To improve ESN learning, \cite{WANG2025130084} introduced a non-convex penalty during the training phase, which helped mitigate overfitting and enhanced generalization. Their approach was tested on four benchmarks, including the Lorenz-63 attractor, and was shown to outperform other ESN architectures. Moreover, ESNs have become widely adopted for chaotic time series analysis. As demonstrated in \cite{shahi_prediction_2022}, multiple ESN architectures were validated across several chaotic benchmarks, often achieving forecasting performance similar to or better than fully trained LSTM and GRU models.

With the emergence of quantum computing, interest is beginning to shift toward extending RC methods to quantum systems, leading to the development of QRC, a new framework that allows to take advantage of the complex dynamics of quantum systems to enhance the representational power of RC \cite{fujii2017harnessing}.  Moreover, several theoretical studies have validated the effectiveness of QRC \cite{tran2020higherorderquantumreservoircomputing,gotting_exploring_2023}, although most have focused on univariate time series due to their reduced dimensional complexity. For example, \cite{suzuki2022natural} proposed a QRC implementation on a real superconducting quantum computer, evaluated on the NARMA task, a benchmark designed to test non-linear memory capacity using delayed feedback systems. Their QRC design is based on input-dependent two-qubit circuits where the Echo State Property (ESP) and system convergence are achieved thanks to the intrinsic quantum noise of hardware. Nevertheless, a key challenge persists: the circuit depth grows with the time series length, posing a limitation on scalability for longer input sequences. To address this issue, the authors in \cite{kobayashi_feedback-driven_2024} proposed reinjecting the previous quantum state along with the current input to encode memory. This method, tested on Mackey-Glass (MG) time series, demonstrated that QRC indeed retains memory, effectively reducing the required circuit length. However, these experiments were conducted through numerical simulations rather than on real quantum hardware.

Although QRC has shown strong performance in univariate time series across various domains \cite{mlika2023user}, its application to MTS remains underexplored. For instance, the authors of \cite{xia_configured_2023} applied QRC to MTS forecasting by encoding two features into a single qubit, one in phase and the other in amplitude. Their method was tested on datasets from biology (gene regulatory networks), finance (foreign exchange market), and nonlinear dynamics (fractional-order Chua’s circuit). A genetic algorithm was used to tune QRC parameters across tasks. Despite the promising outcomes, these results were also obtained through numerical simulation. Similarly focused on MTS prediction with QRC, the authors in  \cite{steinegger2025predicting} proposed a four-qubit Hamiltonian-based QRC to forecast three-dimensional chaotic systems. They leveraged temporal multiplexing by measuring the reservoir multiple times during each evolution period, spatial multiplexing by running several independent quantum reservoirs in parallel, and finally applying a higher-order polynomial transformation to the measurement outcomes to expand the feature space. While this approach considerably decreases the number of qubits, it heavily relies on higher polynomial feature expansion rather than real quantum reservoir features, making it difficult to identify the source of the advantage.

In \cite{ahmed2405prediction} and \cite{ahmed2025optimal}, a QRC model was proposed wherein MTS data is encoded through an input map into a fixed-parameter quantum circuit. Rather than reinjecting past reservoir states, memory is created classically by recursively computing each reservoir state vector using measurements from the quantum circuit and the previous classically computed reservoir state. This approach can be considered as a quantum-enhanced ESN, because the memory is not directly inferred from the quantum system. Similarly, \cite{wang2024application} introduced a circuit-based Extreme Learning Machine to predict elevator Quality of Service—a multivariable task. Features were encoded using rotation gates around the X-axis (R\textsubscript{x}), with one feature per qubit, followed by circular controlled gates. While the implementation drew inspiration from Hamiltonian-based evolution akin to QRC, the dataset did not exhibit time correlation, and the benchmark was conducted on  private data.

Despite the value of numerical simulations, these studies often fail to account for the constraints of real quantum hardware. Moreover, to our knowledge, no study has thoroughly investigated the influence of hyperparameters in QRC from a circuit-based perspective with the intent to deploy on current quantum machines and pursue near-term advantages on MTS.

\section{Background}
\label{sec:background}
\subsection{Reservoir Computing}
\begin{figure}
    \centering
    \includegraphics[width=0.75\linewidth]{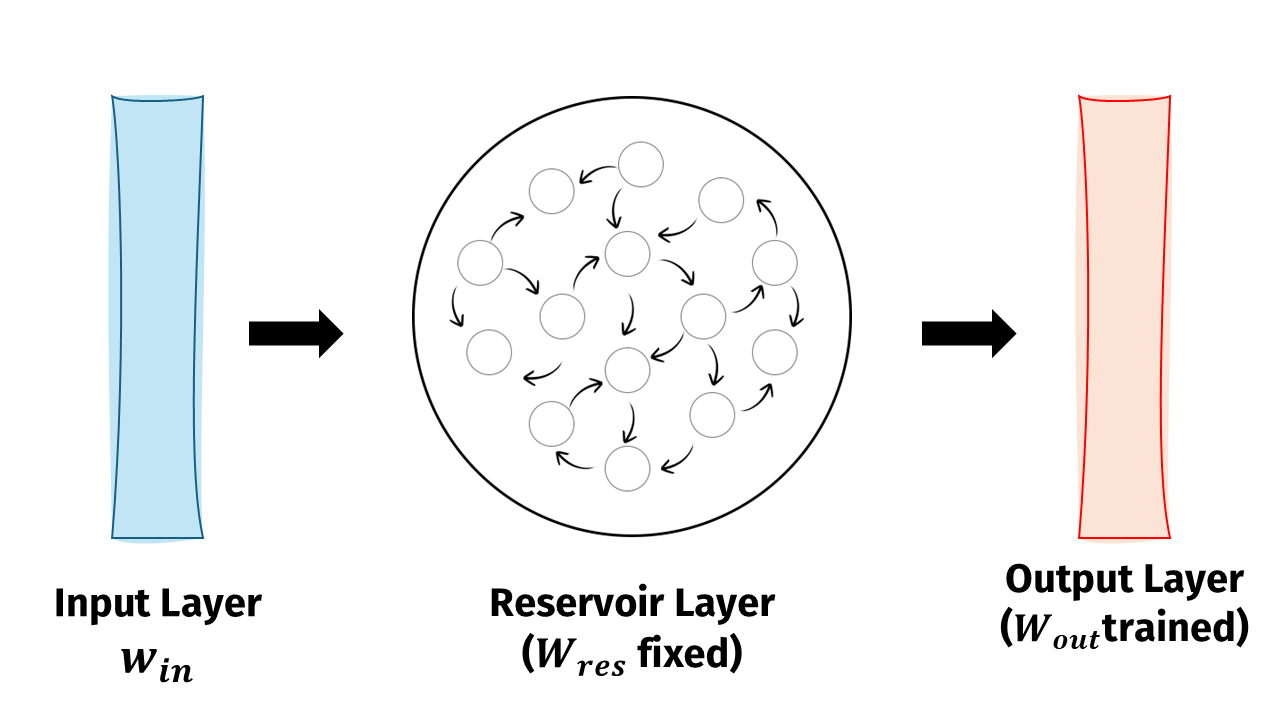}
    \caption{\textbf{Diagram of the RC framework}: Fixed random reservoir with trained output weights.}
    \label{fig:reservoir_diag}
\end{figure}

RC is a general supervised machine learning framework composed of three main components: an input layer, a reservoir layer, and an output layer, as illustrated in Fig. \ref{fig:reservoir_diag}. A defining characteristic of RC is the use of a dynamical system, the reservoir layer, to generate a rich latent representation of the input. This transformation simplifies the role of the output layer, allowing the framework to achieve strong performance using simple output functions.
In practice, the reservoir layer can be computational-based, such as feedforward NNs and RNNs, or physical-based, such as mechanical oscillators, fluid dynamics, or quantum systems~\cite{liang2024physical,Yan2024Emerging}. 

A necessary requirement for the reservoir is to satisfy the Echo State Property (ESP), which reflects its ability to retain short-term memory and produce outputs mostly influenced by recent inputs. In other words, the reservoir should gradually diminish the effect of past inputs on its current state~\cite{10.5555/2770422.2770455}. This fading memory characteristic is quantified by the washout time, which indicates the number of steps over which the reservoir loses dependence on its initial state. An illustrative example of RC is the ESN, where the reservoir is implemented by a fixed, large, and sparsely connected recurrent layer. Unlike traditional RNNs, the weights of the reservoir are not updated through backpropagation, making the training process significantly less computationally demanding. Specifically, the reservoir weight matrix $W_{\text{res}}$ is initialized pseudo-randomly with a spectral radius $\rho(W_{\text{res}}) < 1$.
During inference, the internal reservoir state $r(t)$ is updated at each time step $t$ based on the input $u_t$, as defined by:

\begin{equation}
            r_t = (1- \epsilon)r_{t-1} + \epsilon f(W_{res}r_{t-1} + W_{in}u_t) 
    \label{eq:rc_state_trans}
\end{equation}

\noindent
Here, $W_{\text{in}}$ and $W_{\text{res}}$ are the input and reservoir weight matrices, respectively, $f$ is a nonlinear activation function, and $\epsilon \in [0,1]$ is a smoothing factor that is used to control 
the influence of past reservoir states on the current state.

The final component of the RC architecture is the output layer, which is typically a simple linear regression, where the output $y(t)$ is computed from the reservoir state $r(t)$ as:
\begin{equation}
    y(t) = W_{out} r(t)
    \label{eq:rc_output_eq}
\end{equation}

\noindent
In this formulation, $W_{\text{out}}$ is computed as the closed-form solution to the regularized least squares optimization problem:

\begin{equation}
 W_{out} = \arg \min_W \left\| WR - Y \right\|^2 + \beta \left\| W \right\|^2
 \label{eq:rc_output_opt}
\end{equation}

\noindent
where $R$ stores the reservoir states over the training sequence $\mathbb{U}$, $Y$ is the target sequence and $\beta \in [0,1]$ is a Tikhonov regularization parameter. The final solution is given by
\begin{equation}
 W_{out} = Y R^T (R R^T + \beta I)^{-1}
 \label{eq:rc_output_sol}
\end{equation}
\subsection{Quantum Reservoir Computing}
\label{subsec: QRC}
QRC is a computational framework that adapts the principles of RC to the quantum domain by leveraging the complex dynamics of quantum systems for processing temporal data. Similarly to the classical RC, QRC is formed by an input encoding layer, a dynamic reservoir, and a simple output layer. The reservoir in this case is a quantum system whose intrinsic evolution transforms input signals into high-dimensional quantum states that exhibit rich temporal correlations.  More specifically, for a given MTS, $\mathbb{U} = \{u_{0},u_{1}, \dots, u_{n}, \dots \}$, a QRC protocol proceeds iteratively, applying the following steps to each input $u_n$ in turn:  
\begin{enumerate}
    \item Classical TS encoding: the input $u_{n}$ is injected into the quantum reservoir by applying an encoding unitary  $U_{\mathcal{I}} (u_{n})$ that modifies the state of the system based on that point.

    \item Quantum state evolution: the reservoir undergoes unitary evolution, enabling the locally injected input to propagate through the system via quantum correlations, effectively distributing the information across the entangled many-body state.

    \item Readout: The reservoir state is sampled through repeated quantum measurements yielding expectation values of selected observables $O_k$, such as the Pauli operators.
\end{enumerate}
\noindent
The second step, state evolution, is governed by a unitary $U_{evo}$ which is often selected as a structured unitary derived from a physically motivated Hamiltonian as described in Equation. \ref{eq:ising_evolution}. 
\begin{equation} 
	U_{evo} = e^{-i H \tau} 
\label{eq:ising_evolution}
\end{equation}
Where H is the Hamiltonian representing the underlying physical system’s interactions and  $\tau$ defines the evolution time. However, on gate-based hardware, the continuous-time evolution given by the Equation. \ref{eq:ising_evolution} is not directly available. Therefore, this evolution is approximated using Trotterization, a technique based on the Suzuki-Trotter product formula that allows approximating the exponential of a sum of non-commuting operators by a product of exponentials of individual terms  \cite{Nielsen_Chuang_2010}. Given a Hamiltonian of the form $H = \sum_i H_i$, the first order Trotterized approximation to the unitary time-evolution is
\begin{equation}
e^{-i H \tau} \approx \left( \prod_i e^{-i H_i \tau / \kappa} \right)^{\kappa}
\label{eq:suzuki}
\end{equation} 
where $\kappa$ denotes the number of repetitions. \\ 
Following state evolution, the third step involves leveraging projective quantum measurements to probe the new reservoir state. However, inherently, these measurements induce state collapse, which, in the context of QRC, destroys the memory encoded in quantum correlations, thus hindering temporal information processing.
To mitigate this, various protocols have been proposed \cite{mujal_time-series_2023}. The Restarting protocol addresses the issue by reinitializing the system and repeating steps 1 and 2 for each time step $t \in \{t_0, \dots, t_{n} \}$ before performing the measurements. This preserves the QRC memory of past inputs but increases computational cost due to repeated evolution. Alternatively, the Rewinding protocol leverages the ESP to limit the number of past steps required to the last $\tau_{\text{washout}}$, significantly reducing resource overhead. Other approaches include mid-circuit measurements~\cite{hu_overcoming_2023} and weak measurements~\cite{mujal_time-series_2023}. While the former can be implemented on current hardware, the latter is still a theoretical concept yet to be validated.  In this paper, we primarily concentrate on the Rewinding protocol for practical purposes.

\section{System model}
\label{sec:system_model}
Let $ \mathbb{U} = {\{u_t \in \mathbb{R}^d\}_{t=1}^{T}}$ be a multivariate time series. The goal of our proposed framework is to perform  next-step forecasting using QRC. 
Our target $y$ is, thus, defined as: 
\begin{equation}
y(t )= f_{QRC}(u_{t-1})\approx u_t  
\label{eq:prob_formulation}
\end{equation}
Here $f_{QRC}$ is a QRC-based predictor that evolves the quantum system based on the input $u_{t-1}$ and generates a readout used to estimate $u_{t}$.
\noindent
To achieve this, we propose the Multivariate Time Series QRC (MTS-QRC), a gate-based architecture optimized for real hardware implementation.

\begin{figure*}[!ht]
    \centering
    \includegraphics[width = \linewidth]{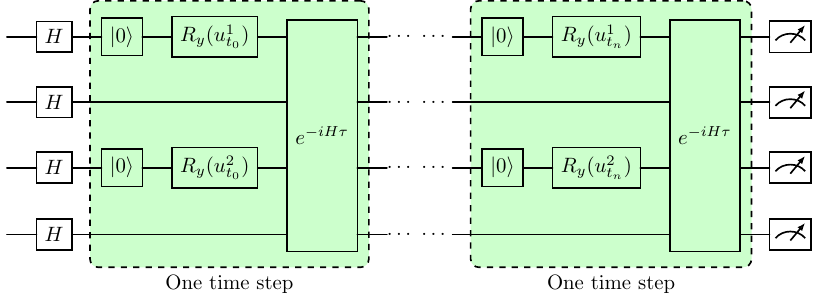}
    \caption{\textbf{MTS-QRC circuit:} Each time series point is input through a circuit block. Within each block, injection qubits are reinitialized and rotated. A Trotterization circuit ($e^{-iH\tau}$) is then applied. The number of blocks reflects the amount of past-point memory desired.}
    \label{fig:mqrc}
\end{figure*}

The reservoir in our model, as being illustrated in Fig.~\ref{fig:mqrc}, is implemented using a quantum circuit composed of $N$ qubits, partitioned into $d$ injection qubits and $m$ memory qubits, with $m \geq d$. Each injection qubit is assigned at least one dedicated memory qubit to retain temporal information. Given the limited qubit-qubit connectivity in real gate based quantum hardware, the qubits are arranged in an alternating pattern: each injection qubit is followed by its associated memory qubits. This layout facilitates interactions between injection and memory qubits, enabling the reservoir to capture temporal dependencies across time steps.

Before injecting the time series, the state of the reservoir is initialized to uniform superposition state. Subsequently, for each time step $u_t \in \mathbb{U}$, a sub-circuit is appended to build the full reservoir circuit as illustrated in Fig.~\ref{fig:mqrc}.\\
\noindent
These sub-circuits are responsible for performing the first and second steps of the QRC protocol outlined in \ref{subsec: QRC}.
Within each one, the classical data encoding step consists in reintializing the injection qubits and applying the encoding unitary $U_{\mathcal{I}}$. Concretely, for each injection qubit $i$, the input feature $u_t^i \in [ 0 ,1]$  is mapped to the state: 
\begin{equation}
	\ket{\psi_i} = \sqrt{1-u_t^i}\ket{0} + \sqrt{u_t^i}\ket{1}  
	\label{eq:qubit_state}
\end{equation}
This mapping is implemented by applying a single-qubit rotation around the $y$-axis: 
\begin{equation}
	R_y(\theta_t^i) \ket{0}, \quad \text{with } \quad \theta_t^i = 2 \arcsin(\sqrt{u_t^i})
	\label{eq:theta}
\end{equation}
Finally, the encoding unitary $U_{\mathcal{I}}$ is given by the tensor product of these rotations. 
\begin{equation}
    U_{\mathcal{I}}(u_t) =   \bigotimes_{i=1}^d R_y(\theta_t^i) 
    \label{eq:encoding}
\end{equation} 
Following data encoding, the system evolves according to the Transverse Field Ising (TFI) Hamiltonian, a widely adopted model in QRC studies. The Hamiltonian is expressed as
\begin{equation}
H_{TFI} = \sum_{\langle i,j \rangle} J_{i,j} X_i X_j + \sum_i h_i Z_i,
\label{eq: ising}
\end{equation}
where $X_i$ and $Z_i$ are Pauli operators acting on spin-$\frac{1}{2}$ particles. The notation $\langle i,j \rangle$ indicates that the summation is restricted to pairs of spins defined as neighbors in the interaction graph; in two dimensions, this corresponds to nearest neighbors on a square lattice, i.e., up, down, left, and right connections. The coupling strengths $J_{i,j}$ modulate the pairwise interaction between spins, while the coefficients $h_i$ control the transverse external field acting on each site. In our case, MTS-QRC considers a spin lattice that corresponds one-to-one with the qubit connectivity of the quantum processor, a model that we call Nearest Neighbors TFI (NN-TFI). This design is suited for real hardware because it respects the hardware connectivity and reduces long-range entangling operations.

\begin{figure}
    \centering
    \includegraphics[width=\linewidth]{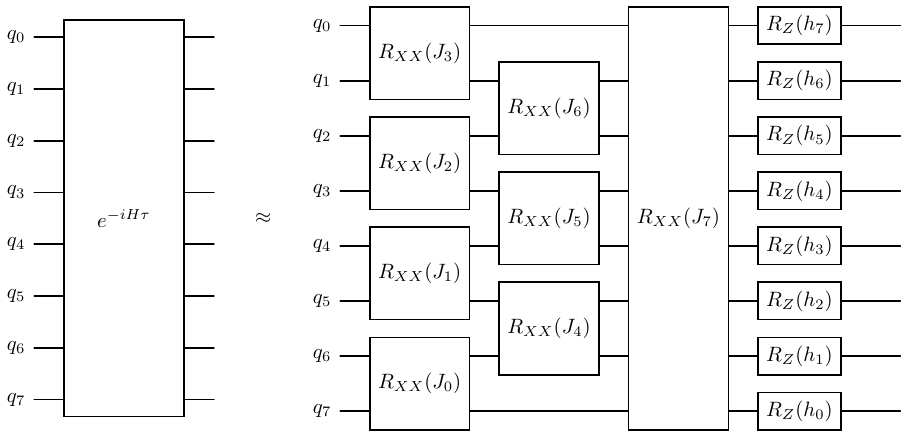}
    \caption{\textbf{Trotterized quantum circuit approximation of time evolution operator $e^{-iH\tau}$corresponding to the Opt-NN-TFI model}. An example with 8 qubits is shown for clarity. The $R_{XX}$ gates are further decomposed into $R_Z$, CNOT, and Hadamard gates. }
    \label{fig:nn_tfi}
\end{figure}

To simulate the corresponding dynamics, we employ Trotterization with evolution time $\tau$ and order $\kappa$. The system’s time evolution can be approximated by a quantum circuit constructed from decomposed TFI Hamiltonian terms. Specifically, the single-qubit terms $e^{-i h_i Z_i \tau / \kappa}$ are implemented using $R_z(2 h_i \tau / \kappa)$ rotations, while the two-qubit interaction terms $e^{-i J_{ij} X_i X_j \tau / \kappa}$ are realized using $R_{XX}(2 J_{ij} \tau / \kappa)$ gates. 
Furthermore, the circuit's gates are arranged as shown in Fig.~\ref{fig:nn_tfi}, in a layout designed to further optimize the  circuit's depth. We refer to this implementation as Optimized NN-TFI (Opt-NN-TFI).\\

Having described each sub-circuit element, we now revisit the complete circuit architecture, Fig.~\ref{fig:mqrc}. The full QRC circuit is constructed by progressively appending sub-circuits that correspond to successive time steps in the input time series. The incremental nature of this construction allows the reservoir’s to accumulate information progressively over time before circuit measurement. The \textbf{Algorithm. \ref{alg:alg1}} details the different steps to extract the reservoir's feature matrix. Specifically, MTS-QRC leverages the Rewind protocol, in which the MTS is processed using a sliding window. This allows the processing of any arbitrarily long MTS with a fixed maximal circuit depth. 

\begin{algorithm}
\KwInput {Time series$\{u_t \in \mathbb{R}^d\}_{t=1}^{T}$, window size $t_w > 1 $, evolution time$\tau > 0$, encoding map$U_\mathcal{I}$, evolution operator$U_{evo}$, number of qubits $N$ } 
\KwOutput {Feature outputs $\{r_t\}_{t=t_w}^{T}$}
\For{$t = t_w$ \KwTo $T$}{
    Initialize reservoir state:$\rho \gets (\ket{+} \bra{+})^{\otimes N}$ \;
    
    \For{$i = t-t_w +1$ \KwTo $t$}
         {Reset injection qubit states: $\rho \gets (\ket{0} \bra{0})^{\otimes d} \otimes \operatorname{Tr}_{\mathcal{I}}(\rho)$\;
         Encode input $u_i$ using map $U_{\mathcal{I}}(u_i)$: $\rho \gets U_{\mathcal{I}}(u_i) \rho U_{\mathcal{I}}^{\dagger}(u_i)$ \;
        
         Apply unitary evolution: $\rho \gets U_{evo}(\tau) \rho U_{evo}^{\dagger}(\tau) $\;}
     Measure reservoir state to obtain features:
     $r_t^{(j)} = \operatorname{Tr}(Z_j\rho)$, for $j = 0, \dots, N-1$\;
}

\Return $\{r_t \in \mathbb{R} ^N\}_{t = t_w }^{T}$ 
\caption{MTS-QRC Feature Extraction }
\label{alg:alg1}
\end{algorithm}

\section{Experiments}
\label{sec:experiments}
\subsection{Datasets}
\subsubsection{Lorenz-63}
The Lorenz-63 system is a simplified dynamical model derived from thermal convection phenomena, where a fluid layer is uniformly heated from below and cooled from above. It captures the essential nonlinear interactions of convective motion in a reduced-order framework.
The differential equations of the system are: 

\begin{figure*}[ht]
    \centering
    \includegraphics[width=0.75\linewidth]{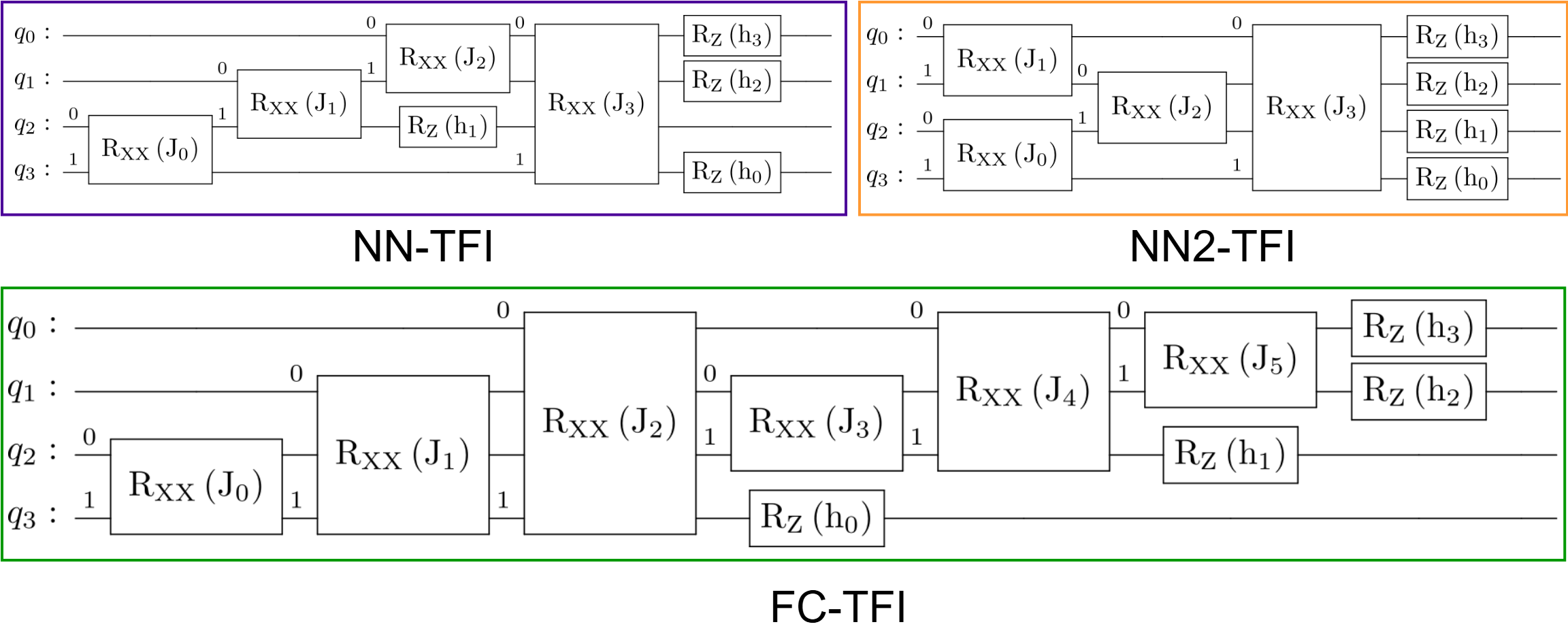}
    \caption{\textbf{Resulting circuits for the Ising Hamiltonian using first-order Trotterization with different qubit connectivity.} FC-TFI represents a fully connected topology where all qubits interact with each other. NN-TFI and Opt-NN-TFI correspond to nearest-neighbor connectivities with different gate layouts.}
    \label{fig:circuits}
\end{figure*}
\begin{equation}
\begin{aligned}
\frac{dx}{dt} &= a(y - x), \\
\frac{dy}{dt} &= x(b - z) - y, \\
\frac{dz}{dt} &= xy - cz
\end{aligned}
\label{eq:lorenz}
\end{equation}
where $(a , b ,c)$ are constants that produce a chaotic behavior if they are fixed at $(10 , 28 , 3/8)$. We generated 1000 data points using the iterative methods of Runge-Kutta, used to approximate solutions of ordinary differential equations (ODEs). The integration was performed with a fixed time step of  $\Delta t = 0.1$ 

\subsubsection{El Niño-Southern Oscillation (ENSO)}
\label{subsec:data} ENSO is a three-dimensional chaotic system used to model the essential physics of ocean-atmosphere interactions in the tropical Pacific. The system variables are $T_e$, $T_w$ and $u$ representing respectively the temperatures of the eastern and western sides of the ocean and the surface wind or zonal current. The differential system is defined as follows: 
\begin{equation}
\begin{aligned}
\frac{du}{dt} &= \frac{B}{2\Delta x}(T_e - T_w) - C(u - u^*), \\
\frac{dT_w}{dt} &= \frac{u}{2\Delta x}(\bar{T} - T_e) - A(T_w - T^*), \\
\frac{dT_e}{dt} &= \frac{u}{2\Delta x}(T_w - \bar{T}) - A(T_e - T^*)
\end{aligned}
\label{eq:enso}
\end{equation}
Here, $B$ quantifies the strength of the current driven by temperature differences, 
$\Delta x$ represents half the ocean width, $C$ models frictional resistance, 
$u^*$ is a reference zonal current, $\bar{T}$ is the deep ocean temperature, 
$A$ governs the rate of heat loss, and $T^*$ is the equilibrium surface temperature. 
The model is parameterized with $B = 940$, $\Delta x = 7.5$, $C = 3$, $u^* = -14.2$, 
$\bar{T} = 16$, $A = 1$, and $T^* = 28$. 
To generate these time series, we used the same approximation method as in Lorenz-63.
\subsection{Experimental Setup}
To validate our approach, we firstly conduct experiments on a noiseless quantum simulator, then we validated the results on quantum hardware.  

MTS-QRC's tunable hyperparameters are summarized in Table~\ref{tab:hyper_summary}.
We consider the number of memory qubits to be 1, 2, 3, and 4 per injection qubit, which means that the total size of the system scales with the TS dimension. In our case, the two considered TS have 3 features each, so the total number of qubits is 6, 9, 12 or 15, respectively.
For the coupling constants $J_{i,j}$, previous studies have sampled values mainly from uniform distributions of range $[0 , 1]$ \cite{gotting_exploring_2023, kutvonen2020optimizing} or from  zero-centered ranges of shape $[-J_e/2 ,J_e/2]$ \cite{steinegger2025predicting,Mart_nez_Pe_a_2021}. In our case, we chose the $[-0.5 , 0.5]$ range because it promotes a richer dynamic via the mixture of ferromagnetic ($J_{i,j} > 0$) and antiferromagnetic ($J_{i,j} < 0$) couplings. The transverse field is set uniformly as $h_i = h$ for all $i \in \{1, \dots, N\}$ with $h = 0.5$. The evolution time $\tau$ is varied to produce both short- and long-time dynamical regimes.

In addition, we compare the performance of our proposed Opt-NN-TFI variant against the Fully Connected TFI (FC-TFI), in which we consider all possible pairs of qubit interaction, and a second version of NN-TFI with different gate layout. An illustrative representation of these layouts is given in Fig.~\ref{fig:circuits}.
From implementation perspective, we apply first-order Trotterization with a single repetition $\kappa = 1$, which balances approximation fidelity and circuit depth \cite {fuchs2024quantumreservoircomputingusing}, and choose the Rewinding protocol with a sliding window of size 10 and step 1 to further reduce the circuit size. As a consequence, the washout time can be considered as $\tau_{w} = 9$. 
The datasets are divided into 80\% training and 20\% test. Each test scenario is run with 5 different seeds for statistical validity.
\begin{table}[ht]
\centering
\caption{Hyperparameter summary.}
\label{tab:hyper_summary}
\begin{tabular}{@{}ll@{}}
\toprule
Parameter & Values \\ \midrule
Number of memory qubits per injection qubits & 1 , 2 , 3, 4\\
Transverse field ($h$)& 0.5 \\
Coupling constant ($J$)& (-0.5 , 0.5) \\
Washout steps ($t_w$)& 9 \\
Evolution time ($\tau$)& 0.01 , 0.1 , 1.0  , 10 \\
Sliding window size & 10 \\ 
Hamiltonian connectivity & FC-TFI , NN-TFI, Opt-NN-TFI \\\bottomrule
\end{tabular}%

\end{table}

Finally, the model's performance is assessed using the Mean Squared Error (MSE), a standard loss metric for regression and forecasting tasks. MSE penalizes larger errors more heavily than smaller ones, making it particularly effective for capturing the fidelity of continuous-valued predictions (see Eq.~\ref{eq:mse}).\\

\begin{equation}
   \mathrm{MSE} = \frac{1}{n} \sum_{i=1}^{n} (y_i - \hat{y}_i)^2 
   \label{eq:mse}
\end{equation}

\section{Results}
\label{sec:results}
In this section, we present a comprehensive evaluation of the proposed method. Our analysis is articulated on four criteria: general performance, comparison with classical benchmarks, validation on the real quantum machine, and study  of the system's hyperparameters.

\subsection{General Results Overview}
We define our baseline as a model that outputs a copy of the input time step $u_t$. This baseline achieves an MSE of 0.0067 on the ENSO series and 0.017 on the Lorenz-63 series. By contrast, our model achieves MSE values of 0.0036 and 0.0087 for ENSO and Lorenz-63, respectively. These results highlight our model's ability to capture the dynamics of both TS and deliver competitive forecasting accuracy. The reported outcomes correspond to the average MSE obtained using the hyperparameter configurations detailed in the Table.~\ref{tab:best_config}. 

\begin{table}[]
\centering
\caption{Best performing reservoir configurations}
\label{tab:best_config}
\begin{tabularx}{0.95\columnwidth}{Xcc}
\toprule
Parameter & ENSO & \multicolumn{1}{c}{Lorenz-63} \\ \midrule
Number of memory qubits & 3 & 3 \\
Transverse field ($h$) & 0.5 & 0.5 \\
Coupling constant ($J$) & (-0.5 , 0.5) & (-0.5, 0.5) \\
Evolution time ($\tau$) & 0.1 & 1.0 \\  \midrule
MSE &0.0036 &0.0087 \\ \bottomrule
\end{tabularx}
\end{table}

\begin{figure}
    \centering
    \includegraphics[width=\linewidth]{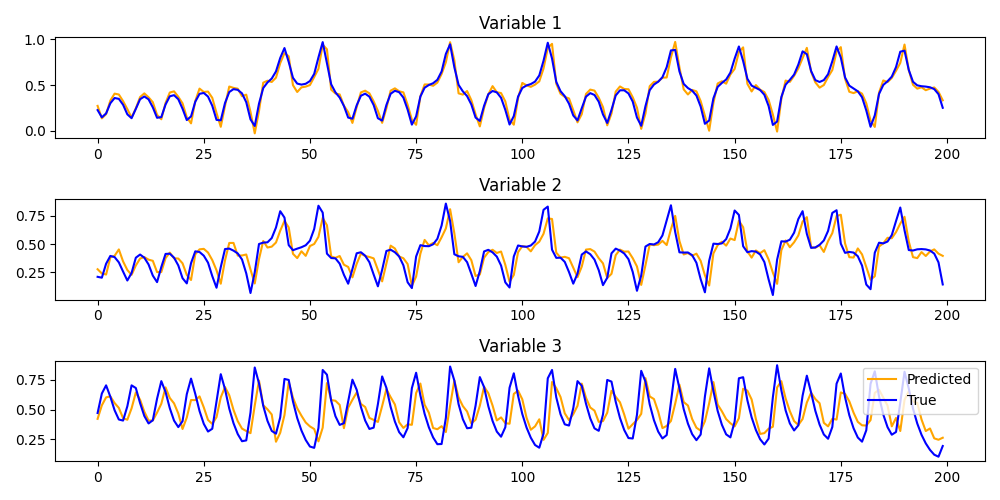}
    \caption{\textbf{Lorenz-63 test set and its corresponding prediction.}}
    \label{fig:Lorenz_test}
\end{figure}

\begin{figure}
    \centering
    \includegraphics[width=\linewidth]{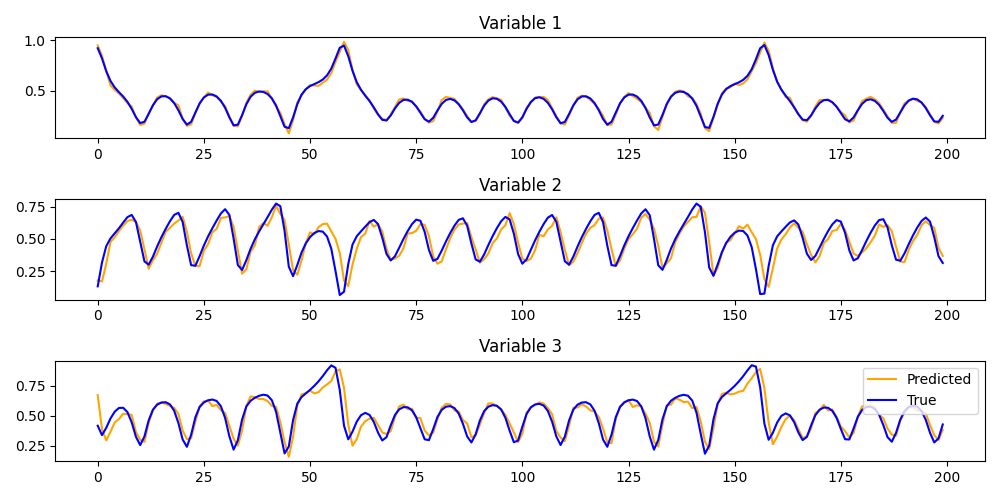}
    \caption{\textbf{ENSO test set and its corresponding prediction}}
    \label{fig:ENSO_test}
\end{figure}

To provide further insight, we illustrate the test TS alongside the corresponding MTS-QRC prediction for the two datasets in the Figs.~\ref{fig:Lorenz_test} and ~\ref{fig:ENSO_test}. Overall, these visualizations confirm the model's capacity to capture the underlying dynamics of each TS. In the case of Lorenz-63, the variable $X$ is predicted with the highest accuracy, achieving an MSE of 0.003, compared to 0.010 and 0.009 for the other two variables. A similar pattern emerges for ENSO where the variable $u$ attains an MSE of 0.0003, while the remaining variables have errors around 0.005 each. In fact, based on the underlying differential systems, variables $X$ (in Lorenz-63) and $u$ (in ENSO) have only linear terms with the other variables in their respective systems, whereas ($Y$,$Z$) and ($T_e$,$T_w$) have rather multiplicative terms each. The consistent behavior of the model suggests that the it learns more effectively the variables with primarily linear correlations, $X$ for Lorenz-63 and $u$ for ENSO, whereas variables with nonlinear dependencies present are more challenging for accurate prediction.


In addition, the visualizations indicate that the model achieves better predictive performance on ENSO than on Lorenz-63, which is consistent with the fact that ENSO exhibits less chaotic behavior. Notably, ENSO is characterized by a low Largest Lyapunov Exponent (LLE), ranging from 0.05 to 0.1 \cite{rojo2023nonlinear}, significantly lower than the value of approximately 0.9 observed in the Lorenz-63 system. This discrepancy in LLE values, which reflects the degree of chaotic behavior, is a probable factor in the differences in model performance between the two systems.

Beyond the baseline comparison, we validate our model by benchmarking it against the classical approaches presented in \cite{shahi_prediction_2022}. This study covers chaotic TS forecasting using RNNs, namely Long-Short Term Memory networks (LSTM), Gated Recurrent networks (GRU), three variants of ESNs and the Nonlinear Vector Auto Regression model (NVAR). The NVAR is a model that constructs nonlinear state vectors using using time-delayed inputs and polynomial expansions to capture nonlinear dependencies within the time series. For a fair comparison, we aimed to replicate the authors' data generation process as closely as possible. However, simulating long time series is computationally consuming on simulator and is infeasible on real hardware. Therefore, we adopted a similar train/test split and ensured that our generated datasets had a comparable number of Lyapunov Times (LT). 
For each model, the authors explored different model sizes and hyperparameters, but we report only the best performance for each. The Table. \ref{tab:realbenchmarks} summarizes the Root Mean Squared Error (RMSE), extracted as accurately as possible from the plots in \cite{shahi_prediction_2022}.
MTS-QRC outperforms LSTM/GRU on both datasets and approaches ESN on Lorenz-63, whereas NVAR (Lorenz) and clustered ESN (ENSO) remain strongest in this setup. For the Lorenz-63 series, our model achieves lower RMSE than the ESNs, though it is outperformed by NVAR. In the case of ENSO, the best-performing method is the Clustered ESN; here, our model surpasses the Hybrid ESN but falls behind the other two variants. Across both series, MTS-QRC consistently outperforms the fully learnable models, namely LSTM and GRU.

\begin{table}[ht]
\centering
\caption{RMSE Classical approach comparison}
\label{tab:realbenchmarks}
\begin{tabularx}{0.95\columnwidth}{Xcc}
\toprule
Model & Lorenz-63 & ENSO \\ \toprule
Baseline & 0.130 & 0.081 \\ \midrule
LSTM & 0.25 & 0.200 \\
GRU & 0.28 & 0.200 \\  \midrule
ESN & 0.168 & 0.018 \\
Clustered ESN & 0.178 & \textbf{0.007} \\
Hybrid ESN & 0.178 & 0.074 \\ \midrule
NVAR & \textbf{0.042} & \textbf{0.014} \\ \midrule
MTS-QRC (Best) & \textbf{0.093} & 0.060 \\ \bottomrule
\end{tabularx}
\end{table}

\subsection{Hardware Results }
To evaluate the practical viability of the proposed approach beyond simulation, we run the model on IBM Quebec's Heron R2 quantum processor. It is a 156 superconducting qubit chip supporting up to 5000 gate operations.
Particularly, we run the hyperparameter configuration presented in Table. \ref{tab:best_config} with only one seed for each dataset. Given that we are using sliding windows of 10 points, the considered configurations produce 990 circuits per TS. For maximum circuit efficiency, we apply Qiskit's highest optimization level. As a consequence, these  circuits have approximately 254 layers and 1858 gates each. Here, a layer refers to a set of quantum gates that can be applied in parallel, such that no two gates are acting on the same qubit. Using the default shot number set to 4096, the 990 circuits can be executed on the Quantum Processing Unit (QPU) in approximately 20 minutes for each TS.\\

\begin{figure}[ht]
    \centering
    \includegraphics[width=0.75\linewidth]{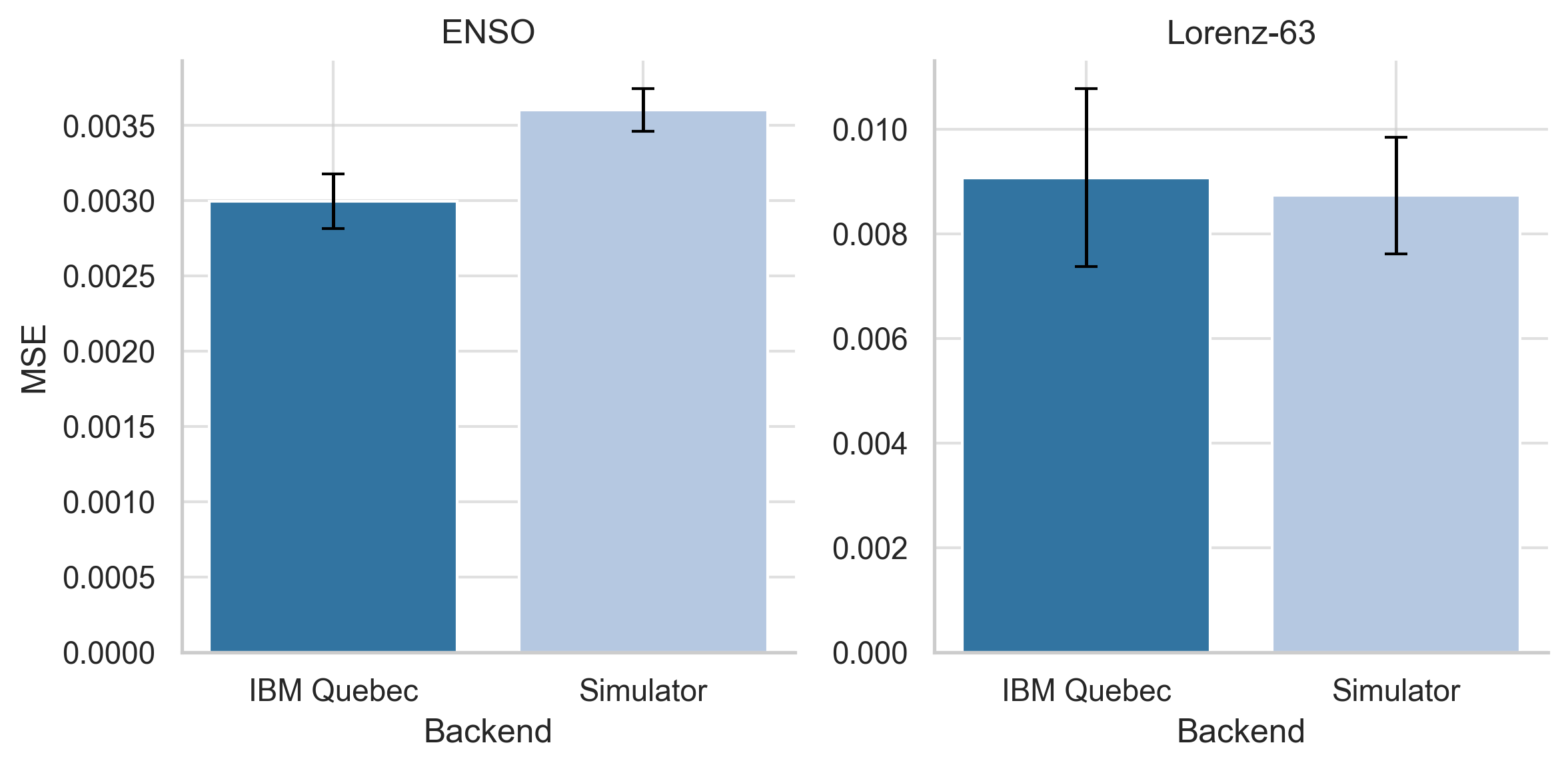}
    \caption{\textbf{Performance of real quantum hardware across datasets.} For the Lorenz–63 system, results obtained on real devices closely match those from noiseless simulations, while for the ENSO dataset, hardware performance even surpasses the noiseless baseline.}
    \label{fig:Harrin_res}
\end{figure}

The Fig.~\ref{fig:Harrin_res} illustrates the achieved results alongside those of a noiseless simulator. The MTS-QRC model maintains a lower MSE than the baseline on both datasets, demonstrating that the model approach is viable on real hardware. However, we observe a surprising trend in these results: on the ENSO dataset, the forecasting MSE obtained from noisy real hardware executions is lower than that of the noiseless simulator. This counterintuitive behavior does not appear for the Lorenz-63 dataset, where hardware noise yields the expected performance deterioration.\\

To better quantify this effect, we performed three independent hardware runs of the ENSO experiment and compared them against five simulator runs of the same configuration. The resulting averages and standard deviations are reported in Fig.~\ref{fig:Harrin_res}. For ENSO, the hardware average MSE is 0.0030, which remains lower than the simulator average reported in Table \ref{tab:best_config}. In contrast, for the Lorenz-63 system, the hardware runs mostly yield higher errors than the simulator, confirming that the unexpected improvement is specific to ENSO.

Since the final forecasting is obtained through a linear regression function, we study the Singular Value Decomposition (SVD) of the training features collected from the simulator and the real hardware to identify the reason behind the performance differences. SVD provides mathematical insights into the intrinsic structure of the feature matrices, highlighting their effective dimensionality and variance distribution \cite{mandel1982use}.
In practice, we start by standardizing each training reservoir output feature, and we compute the SVD given by 
\begin{equation}
    R_{train} = U diag(\sigma_1, \dots,\sigma_N) V^T 
\end{equation}
Here $U$ and $V$ are orthogonal matrices, the $~\{\sigma_1 >\sigma_2 >  \dots>\sigma_N\}$ are the singular values, and $R_{train}$ contains the reservoir's features over the training period that is obtained by stacking $\{ r_t \in R^N\}_{t=0}^{T_{Train}}$. Since we are studying the particular case of 12 qubits, $N$ here is equal to 12. 

We inspect the singular value spectrum and the explained variance ratio, as given in Equation.~ \ref{eq:var_rat}. These quantities reveal how the information is distributed across different directions in the feature space~\cite{liu2022exploring,en16083572}. Based on it, we understand the impact of the quantum hardware noise on the distribution of variance and thus the stability and effectiveness of linear regression.
\begin{equation}
    \text{explained variance ratio}_i = \frac{\sigma_i^2}{\sum \sigma_i^2} \text{ , } \forall \sigma_i \in \{\sigma_1, \dots,\sigma_N\}
    \label{eq:var_rat}
\end{equation}

\noindent
For Lorenz-63, the effective matrix rank, obtained by Equation.\ref{eq:hhi}, increases from 9 on the simulator to 10 on the hardware, suggesting that information is more dispersed in the hardware features.

\begin{equation}
\text{Rank}_{\text{eff}}(A) = \frac{1}{\text{HHI}} 
= \frac{1}{\sum_{i=1}^n p_i^2}, 
\qquad p_i = \frac{\sigma_i}{\sum_{j=1}^n \sigma_j},
\label{eq:hhi}
\end{equation}

\begin{figure}
        \centering
        \includegraphics[width=0.75\linewidth]{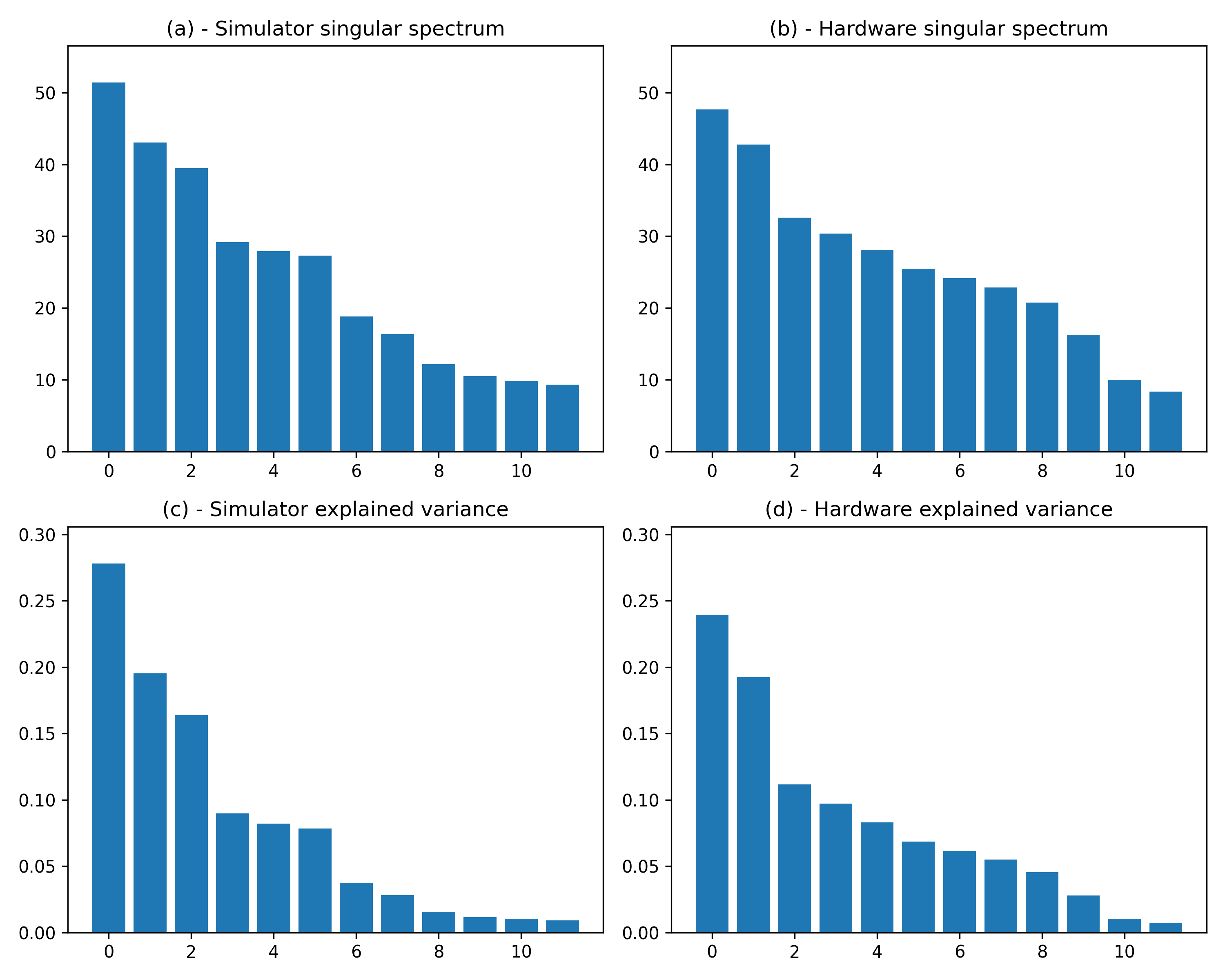}
        \caption{\textbf{Lorenz-63: SVD Metrics.} Panels (a) and (b) show the singular value spectra for the simulator and hardware, respectively. Panels (c) and (d) display the corresponding explained variance ratios for the simulator and hardware.}
        \label{fig:lorenz-svd}
\end{figure}

From the singular value spectrum, we observe in Fig.~\ref{fig:lorenz-svd}-a and b that the largest singular value shrinks on the hardware and the singular spectrum becomes flatter. This behavior similarly appears in the explained variance ratio in  Figs.~\ref{fig:lorenz-svd}-c and ~\ref{fig:lorenz-svd}-d: the simulator features concentrate variance strongly in the leading direction, whereas hardware features show a flatter distribution across directions, especially starting from direction 3.
This shift in the distribution explains why forecasting accuracy is higher on simulator data: when variance is concentrated in fewer dominant directions, linear regression can more reliably capture the underlying dynamics with fewer effective parameters. In contrast, the hardware noise seems to amplify other less informative directions, diluting the information and reducing regression reliability. In this case, hardware noise acts destructively in Lorenz-63 by dispersing the feature variance.

\begin{figure}
    \centering
    \includegraphics[width=\linewidth]{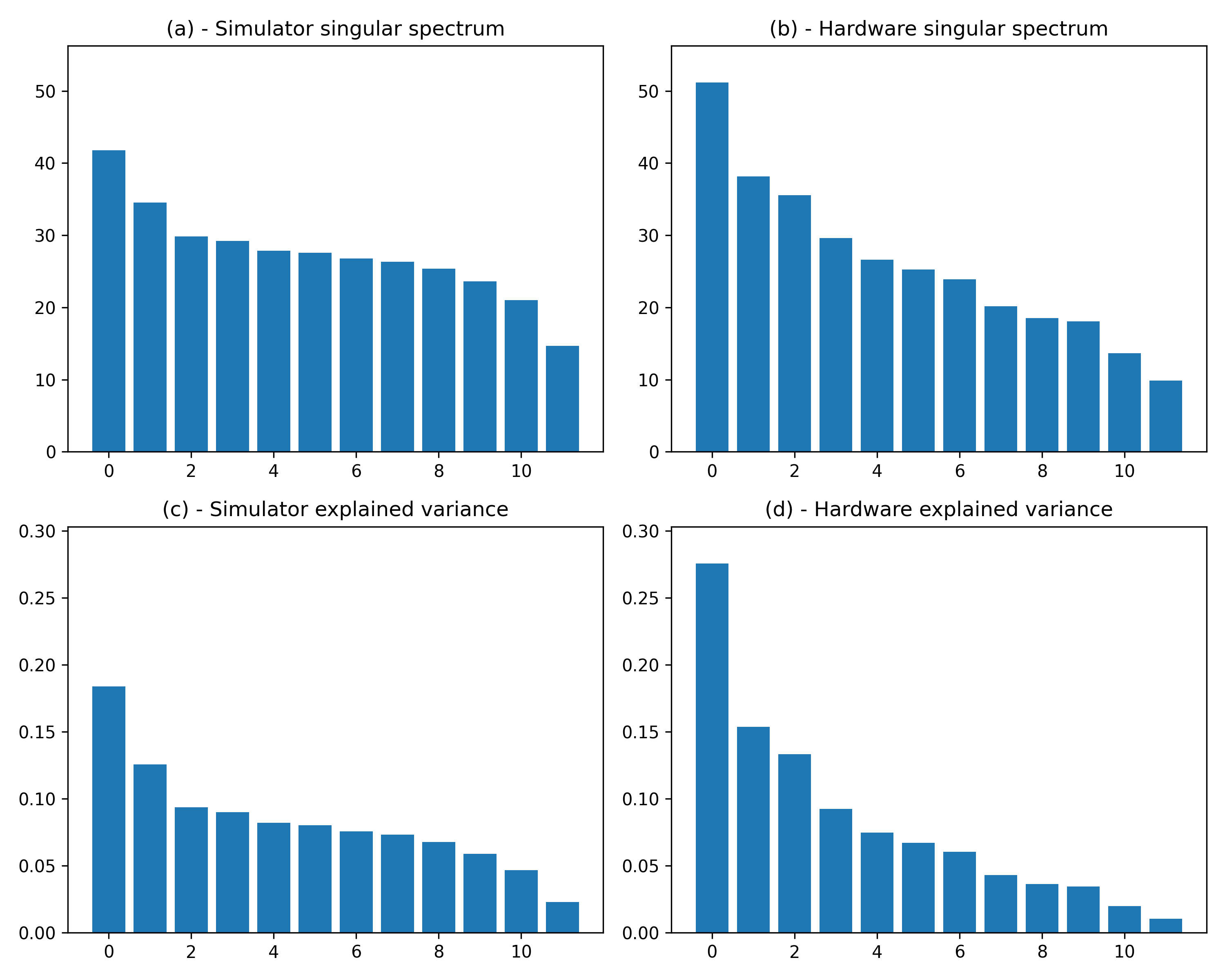}
    \caption{\textbf{ENSO: SVD metrics.} (a) and (b) show the singular value spectra for the simulator and hardware, respectively. (c) and (d) display the corresponding explained variance ratios.}
 \label{fig:enso_svd}
\end{figure}

In contrast, for ENSO, the effective matrix rank decreases from 11 on the simulator to 10 on the hardware, suggesting that the feature space becomes more compact. The singular value spectrum, illustrated in Fig.~\ref{fig:enso_svd}-a and ~\ref{fig:enso_svd}-b, shows that simulator features are relatively flatter, while hardware features emphasize one dominant singular value more strongly. Consistently, the explained variance ratio, Fig.~\ref{fig:enso_svd}-c and ~\ref{fig:enso_svd}-d, indicates that simulator features spread variance across multiple directions, while hardware features concentrate variance more sharply into a leading direction with the others significantly reduced.
This compaction explains why forecasting improves on hardware features in the ENSO case: the hardware noise, rather than diluting the information, effectively acts as a regularizer, suppressing weaker directions and concentrating predictive power into a smaller number of dominant modes. Thus, hardware noise acts constructively here, enhancing regression performance.
 
Despite the limited scope, our results suggest that, in certain settings, noise can actually be beneficial by acting as a regularizer that improves the numerical conditioning of feature representations.

\subsection{Hyperparameters tests} 
\begin{figure}
    \centering
    \includegraphics[width=0.75\linewidth]{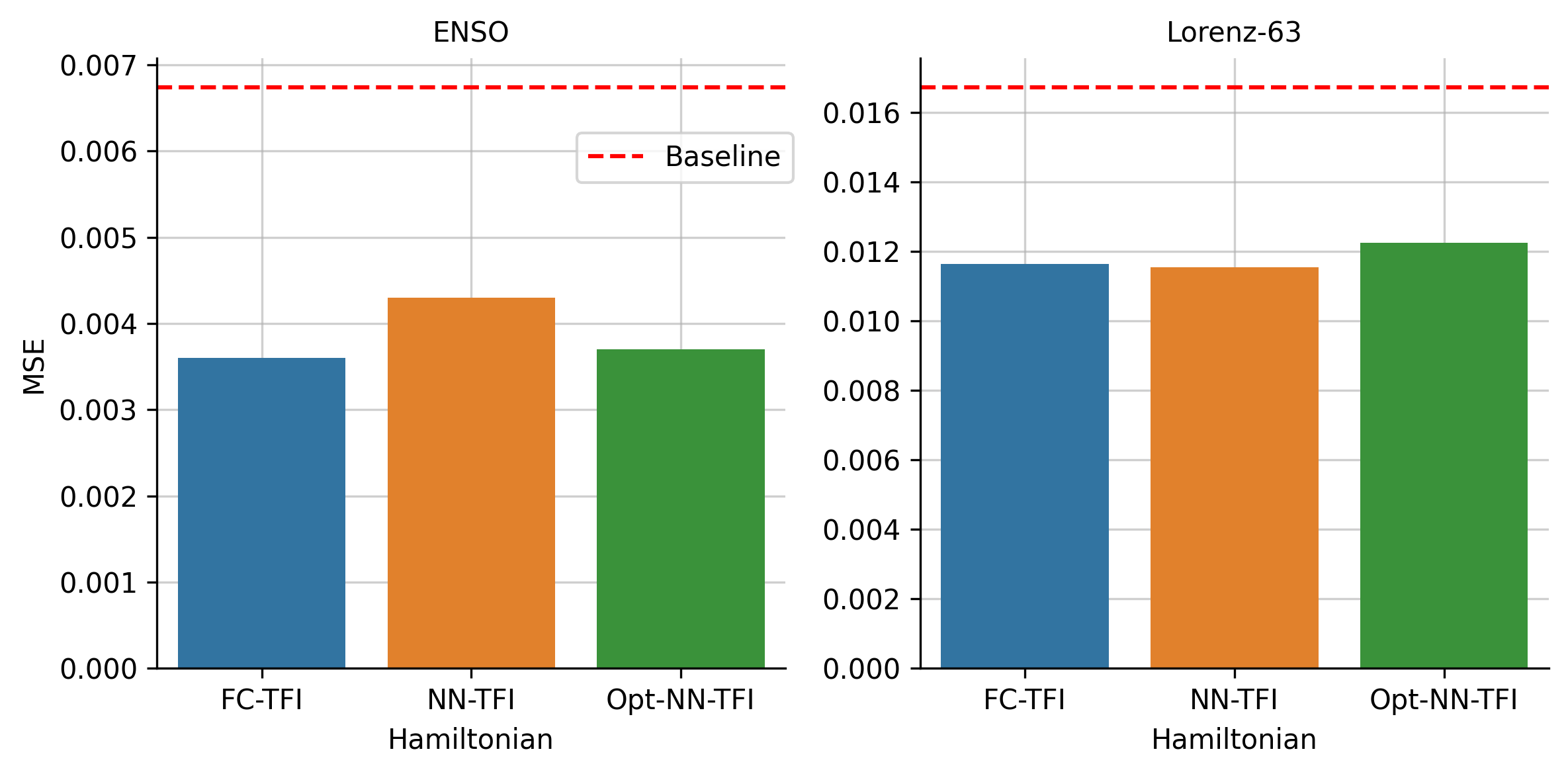}
    \caption{\textbf{MSE comparison of different TFI connectivities (FC-TFI, NN-TFI and Opt-NN-TFI) on two time series datasets (ENSO and Lorenz-63) with 10-point sliding window Rewind protocol}. Less performance variance is observed across the connectivity configurations. Additionally, Opt-NN-TFI performs comparably to the other two, enhancing its applicability on real hardware.}
    \label{fig:ham_ts}
\end{figure}

Our approach leverages a variant of TFI rarely studied in the context of QRC; therefore, we compare its effectiveness against two baselines: the FC-TFI and another NN-TFI. 
As shown in Fig.~\ref{fig:ham_ts}, our model achieves a comparable performance to FC-TFI and NN-TFI on both TS, though it records a slightly higher MSE, about 5\%, on the ENSO series.

\begin{figure*}[ht]
    \centering
    \includegraphics[width=0.75\linewidth]{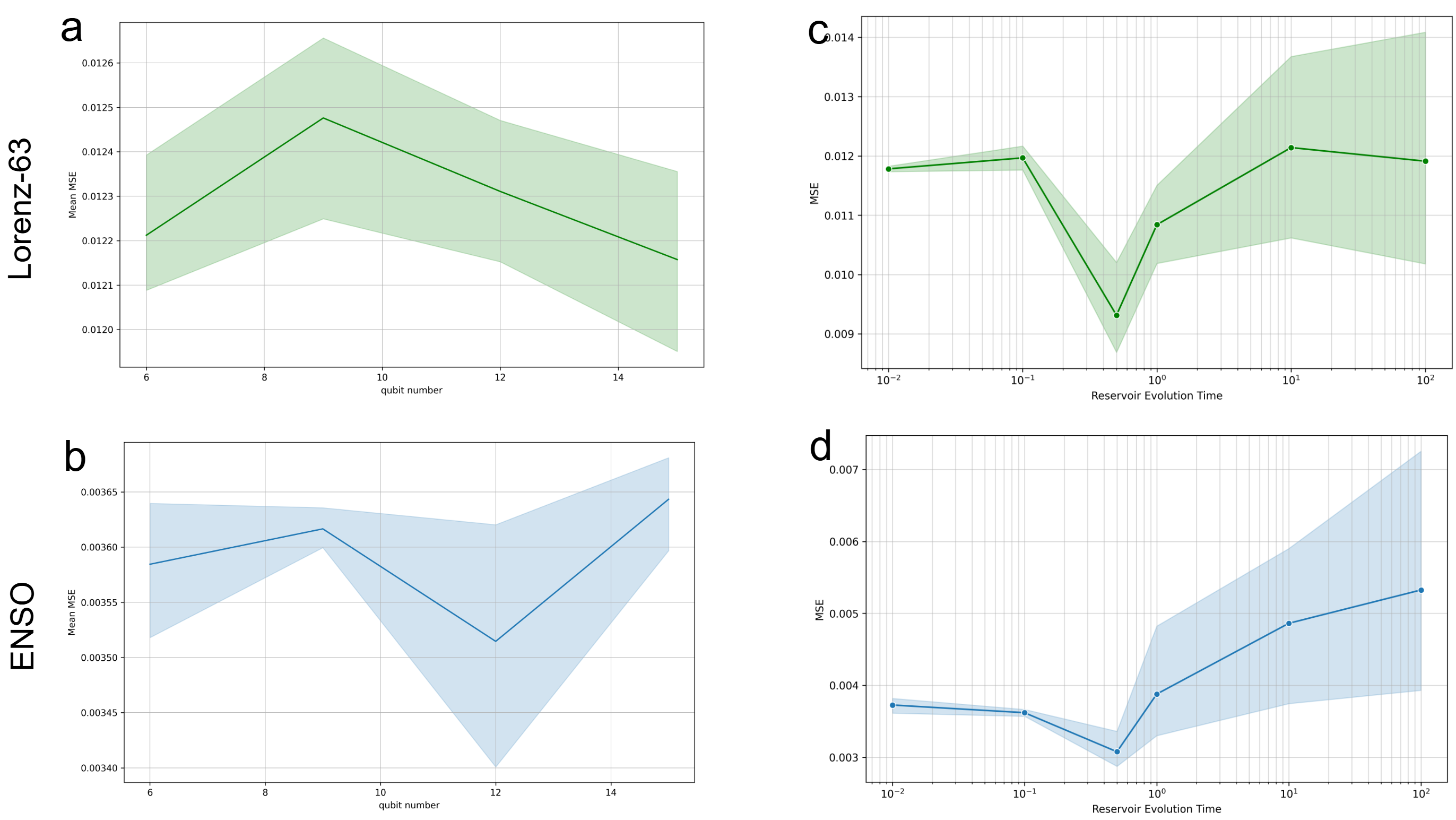}
    \caption{\textbf{Hamiltonian hyperparameter effect on MSE for ENSO and Lorenz-63 datasets.} The experiences are based on the root configuration consisting of 1 injection qubit per memory qubit, $J_{i,j}$ sampled from uniform distribution $[-0.5, 0.5]$, $h = 0.5$ and $\tau = 0.1$. For each plot, we fix all the other hyperparameters and only vary the studied one. }
    \label{fig:hyperl}
\end{figure*}

Since the model design allows a flexible choice of number of memory qubits and given the common practice in univariate TS QRC studies of assigning multiple memory qubits. We examine the MSE across different qubit counts in the case of MTS-QRC.
In Fig.~\ref{fig:hyperl}, subfigures (a) and (b) present the MSE with regard to the number of qubits for Lorenz-63 and ENSO respectively. It is important to recall that, when we increase the qubit number, we are actually increasing the number of memory qubits dedicated to each injection qubit. 
In the case of Lorenz-63 TS, we observe that the MSE peaks at 9 qubits, i.e., 2 memory qubit per injection qubit, and then decreases. The improvement of the average MSE resulting from increasing 6 to 15 qubit is around 1.6\%. In contrast, ENSO exhibits a different pattern, where the average MSE slightly increases passing from 6 to 9 qubits, then decreases at 12 qubits before rising again. Based on the average MSE, 12 qubits appears optimal; however, this configuration exhibits high variability across random seeds, unlike the 9-qubit case, which is more stable. In our experiments, the the coupling constants $\{J_{i,j}\}$ values are the only seed-specific hyperparameter, meaning that in the high-variability scenarios, the specific values of $J$ have a strong influence on performance. Consequently, selecting the optimal number of qubits depends on the average MSE and on the size of the data. In our case we deliberately used low-dimensional systems that allow us to vary the number of qubits.

We also analyze the effect of the evolution time $\tau$. Figs.~\ref{fig:hyperl}-b and ~\ref{fig:hyperl}-d show that the optimal performance is achieved between 0.1 and 1. In particular, increasing from 0.01 to 0.1 causes small MSE changes, slightly increasing for Lorenz-63 and decreasing for ENSO, while values greater than 1 sharply degrade performance. 
Apart from their physical interpretations, in circuit-based implementation of the TFI model, the variables $\tau$, $h$ and $J_{i,j}$ contribute as rotation angles. For instance, $\tau$ and $h$ contribute to $R_z$ rotation applied to all qubits as $\theta = 2 h \tau$. Given the periodicity, the product of these two parameters yields an angle in the range of $[0 , \pi]$ that is high enough to produce meaningful state change but not arbitrarily high, as it deteriorates the performance of the model, a behavior we see for large values of $\tau$ in Fig.~\ref{fig:hyperl}.

\begin{figure}[ht]
    \centering
    \includegraphics[width=0.75\linewidth]{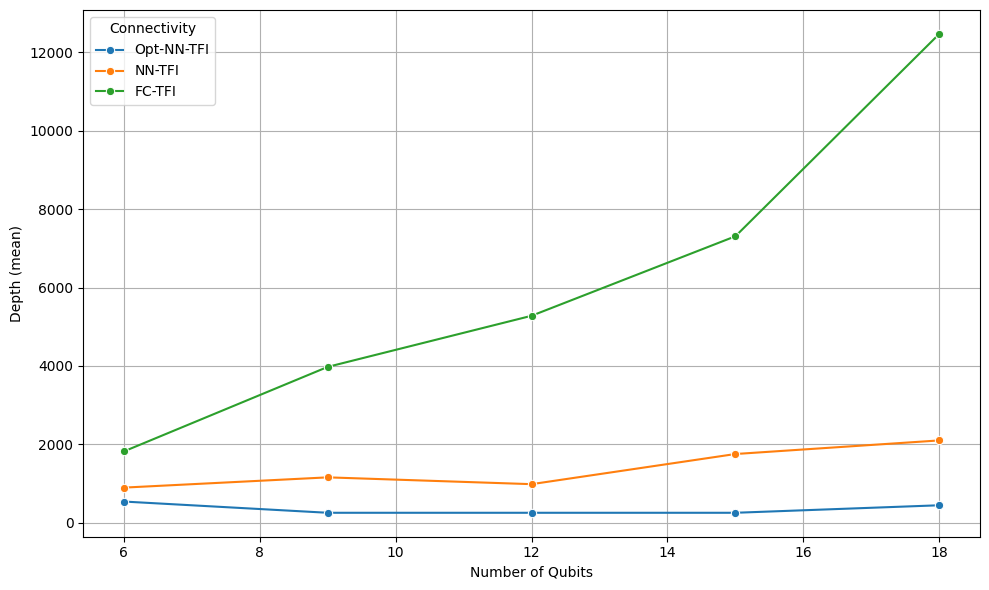}
    \caption{\textbf{Full MTS-QRC circuit depth as a function of the number of qubits for each variant of the evolution unitary $U_{evo}$.} The FC variant's depth scales exponentially with the number of qubits whereas the NN variants' remain almost constant.}
    \label{fig:depth}
\end{figure}

Taken together, the hyperparameter study highlights that optimal parameter choices are strongly dependent on the specific TS dynamics. The differing result trends across series indicate that fully leveraging the potential of QRC requires carefully tailoring the configuration to the data.
Since accuracy alone is insufficient for practical implementation, circuit complexity is critical when deploying QRC on real hardware.
To quantify this, we estimate the required resources to build the full circuit for processing a 10-point time window across different qubit counts. This estimate is obtained using a simulated version of the IBM Quebec hardware employed in our real-hardware experiments. For each configuration, we transpiled the circuits 50 times with different transpiler seeds. As expected, FC-TFI generates circuits significantly deeper than the two NN-TFI and Opt-NN-TFI. For Instance, using 6 qubits FC-TFI generates circuits with $1561 \pm 44$ layers and $3028 \pm 38$ gates, compared to $543 \pm 48$ layers in the case of our model. Moreover, while the circuit depth of FC-TFI shows a rapid increase with the number of qubits, that of our model remains constant, thereby supporting the rationale behind our design choice. Fig.~\ref{fig:depth} further supports this conclusion: the NN-TFI presents a significantly reduced circuit depth, yet it doesn't achieve the desired reduction because of the unoptimized gate layout. Most importantly, our model decouples qubit number from circuit depth. offering more flexibility. 

\section{Conclusion}
\label{sec:conclusion}
In this work, we proposed a gate-based QRC architecture for multivariate time series forecasting. Our approach leverages Hamiltonian-based quantum evolution and a rewind protocol for efficient reservoir information retrieval. Specifically, we adopted trotterize NN-TFI with reduced circuit depth, making it suitable for NISQ-era hardware.
Our proposed model performs competitively with respect to classical RNNs, including various ESN, on the benchmark TS  Lorenz-63 and ENSO. We further tested its feasibility on IBM Heron R2 quantum hardware, where we observed that the quantum noise improved the predictive capacity for the ENSO dataset while maintained close to simulator results on Lorenz-63. This observation provides empirical evidence of a potential positive role for quantum hardware noise in QRC. In addition, we provided an analysis of hyperparameter influence, confirmed our design choices, and provided insights into the architecture’s behavior.
Future works include extending the study to other Hamiltonians and higher-dimensional datasets, producing  deeper quantum circuits to validate the robustness on real devices. 
Additionally, a systematic study to identify the specific operational conditions, parameters, and data structure that ensure a consistent emergence of this enhancement will be conducted to shed light  on how noise can be harnessed as a resource in quantum-enhanced computation.

\section*{Acknowledgment}

The authors would like to extend their gratitude to the Natural Sciences and Engineering Research
Council of Canada, Prompt, Thales Digital Solutions,
and Zetane Systems for their financial support. 



\end{document}